# Still Not There: Can LLMs Outperform Smaller Task-Specific Seq2Seq Models on the Poetry-to-Prose Conversion Task?

**Kunal Kingkar Das, Manoj Balaji Jagadeeshan, Nallani Chakravartula Sahith, Jivnesh Sandhan[†] and Pawan Goyal**

IIT Kharagpur, India and [†]Kyoto University, Japan

{kunalkingkar13, manojbalaji1, ncsahith, pawang.iitk}@gmail.com, jivnesh@i.kyoto-u.ac.jp

## Abstract

Large Language Models (LLMs) are increasingly treated as universal, general-purpose solutions across NLP tasks, particularly in English. But does this assumption hold for low-resource, morphologically rich languages such as Sanskrit? We address this question by comparing instruction-tuned and in-context-prompted LLMs with smaller task-specific encoder–decoder models on the Sanskrit poetry-to-prose conversion task. This task is intrinsically challenging: Sanskrit verse exhibits free word order combined with rigid metrical constraints, and its conversion to canonical prose (anvaya) requires multi-step reasoning involving compound segmentation, dependency resolution, and syntactic linearisation. This makes it an ideal testbed to evaluate whether LLMs can surpass specialised models.

For LLMs, we apply instruction fine-tuning on general-purpose models and design in-context learning templates grounded in Pāṇinian grammar and classical commentary heuristics. For task-specific modelling, we fully fine-tune a ByT5-Sanskrit Seq2Seq model. Our experiments show that domain-specific fine-tuning of ByT5-Sanskrit significantly outperforms all instruction-driven LLM approaches. Human evaluation strongly corroborates this result, with scores exhibiting high correlation with Kendall's Tau scores. Additionally, our prompting strategies provide an alternative to fine-tuning when domain-specific verse corpora are unavailable, and the task-specific Seq2Seq model demonstrates robust generalisation on out-of-domain evaluations. Our code[1] and dataset[2] are publicly available.

## 1 Introduction

LLMs have become the default solution across NLP applications, often replacing specialised ar-chitectures in English (Zhao et al., 2025; Minaee et al., 2025). In contrast, Sanskrit NLP has historically relied on carefully designed linguistic and task-specific systems for segmentation, morphology, compound processing and dependency parsing (Sarkar et al., 2025; Ray et al., 2024; Sandhan et al., 2023a). Given the recent dominance of LLMs in English, a natural question arises: are these specialised Sanskrit NLP efforts still necessary, or should future research increasingly rely on general-purpose LLMs?

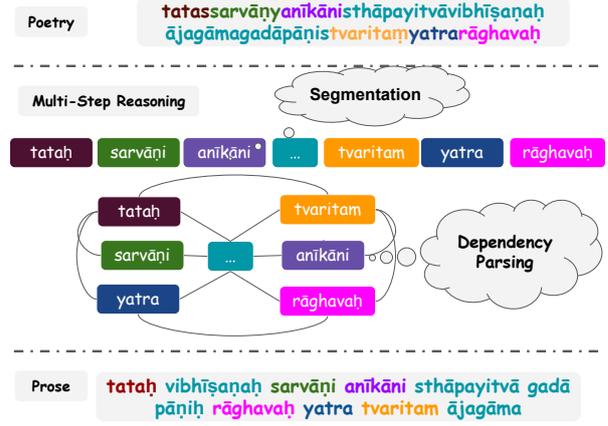

Figure 1: Illustration of the challenges in the poetry-to-prose conversion task. The top panel shows the input verse and the bottom panel shows the corresponding prose output. The middle panels depict two intermediate reasoning steps: segmenting compound/sandhi-merged tokens and performing dependency parsing for the toy example.The color-coded schema highlights sandhi phenomena where word boundaries become merged.

To investigate this question, we focus on the Sanskrit Poetry-to-Prose Conversion task (*Anvaya*), which serves as a natural stress test for evaluating whether LLMs can surpass task-specific models. Figure 1 illustrates that Sanskrit poetry-to-prose conversion is far more challenging than standard word-order linearisation (Kulkarni, 2010; Krishna

---

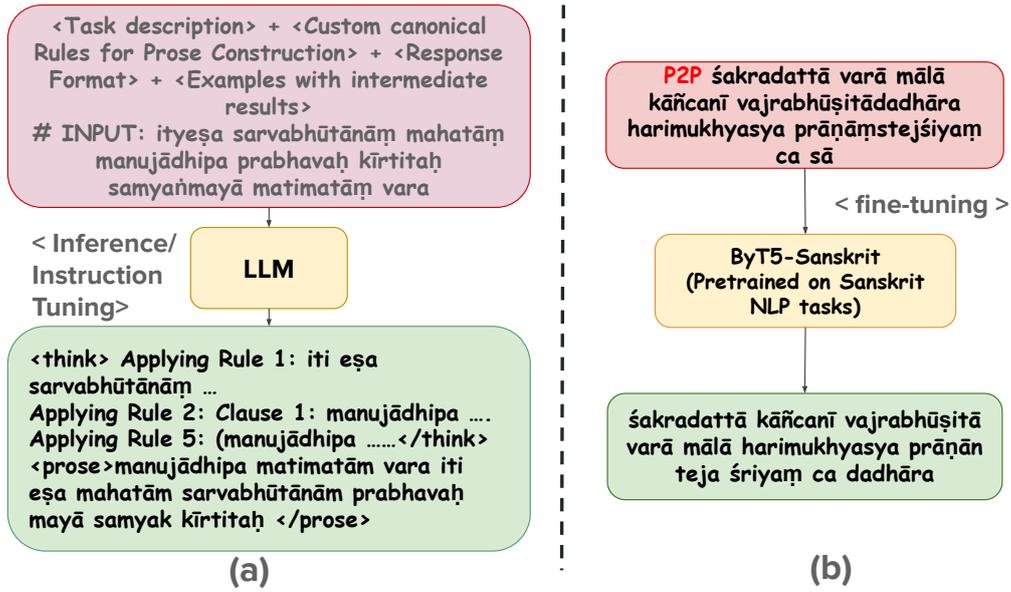

Figure 2: Illustration of the two modeling paradigms using a toy example: (a) LLM (decoder-only): We apply instruction tuning with linguistically informed templates and also evaluate in-context learning. (b) Task-specific Seq2Seq (encoder–decoder): We fine-tune the ByT5-Sanskrit model, already trained on segmentation and dependency-parsing tasks, to endow it with additional poetry-to-prose conversion capability.

et al., 2019).[3] The task requires coordinated multi-step reasoning: segmenting densely fused compounds into valid constituents, inferring syntactic dependencies and linearising the resulting structure into coherent prose. These processes operate jointly rather than sequentially, relying on linguistic principles rooted in Pāṇinian grammar and classical commentary heuristics.

We compare two modelling paradigms as shown in Figure 2. For LLMs, we perform instruction fine-tuning of general-purpose models and construct in-context learning templates explicitly grounded in Pāṇinian principles. For task-specific modelling, we fully fine-tune a ByT5-Sanskrit encoder–decoder model that directly operates on byte-level representations, which are well-suited for low-resource, morphologically rich languages. Our results show that the domain-specific fine-tuning of ByT5-Sanskrit substantially outperforms all instruction-driven LLM approaches. Human evaluation strongly supports this finding, with scores showing high correlation with Kendall's $\tau$ and confirming the limitations of current LLMs on this linguistically intensive task. At the same time, our prompting strategies offer a practical alternative when domain-specific verse corpora are un-available, while the specialised Seq2Seq model exhibits robust generalisation to out-of-domain inputs. These observations collectively demonstrate that, despite recent advances, LLMs are still not capable of surpassing task-specific models on the structurally demanding problem of Sanskrit poetry-to-prose conversion.

## 2 Methodology

We frame the poetry-to-prose conversion as a Seq2seq task that implicitly requires the model to handle the sub-tasks of segmentation, dependency resolution, and reordering. We compare two modeling paradigms: (1) LLMs (Decoder-only) (2) Task-Specific models (Encoder-Decoder)

**Design of Instruction prompts:** The poetry to prose task requires explicit grammatical rules. We use the canonical word-ordering rules of *Daṇḍa-anvaya-janaka* (Kulkarni et al., 2019) (Appendix D) to provide a foundational framework, but they are not comprehensive; constructions involving multiple clauses, appositives, or embedded phrases fall outside their scope. Encoding every such edge case directly into a prompt is impractical, as it would inflate the prompt and exhaust the model's context window. To balance coverage and efficiency, we design a logical, step-by-step prompt inspired by Wei et al. (2022). Our approach integrates custom-modified rules with intermediate reasoning steps in a Chain-of-Thought (CoT) format (Appendix B). To evaluate the impact of these design choices, we conduct a prompt ablation study using the Phi-4 model (Appendix E). As shown in Table 6, incorporating linguistic rules ($P_{full}$) yields a substantial performance gain over a basic task description ($P_{base}$), nearly doubling the BLEU score. Moreover, structuring the reasoning explicitly in CoT format provides an additional 15.3% improvement over the full prompt. These results confirm that structured, domain-aware prompting is essential for effectively leveraging general-purpose LLMs on this linguistically demanding task.

---

[3]Translation: Then Vibhīṣaṇa, having stationed all the armies, swiftly came, mace in hand, to the place where Rāghava (Rāma) was.

### 2.1 LLMs (Decoder-only)

We treat the problem as a multi-step process, guiding these generalist models with explicit, linguistically-grounded instructions. This is reflected in our prompt design for IFT and ICL, which breaks down the task into continuous steps.

**Instruction Fine-Tuning (IFT):** We evaluate whether providing explicit linguistic rules can guide models with broad capabilities, such as LLMs, toward this specialized objective, (see Appendix G). We use models from the Llama-3.x-Instruct (1B, 3B, 8B) (Meta AI, 2024a,b) , Phi-4 (14B) (Abdin et al., 2024) and, Qwen-2.5 (7B) (Qwen et al., 2025) series. The finetuning prompt is curated based on the default instruction prompt of the official models and the Anvaya rules from the canonical word-ordering rules of *Daṇḍa-anvaya-janaka* (Kulkarni et al., 2019). The complete instruction template is provided in Appendix B.

**In-context Learning (ICL):** The In-context learning (Brown et al., 2020) is a paradigm that allows LLMs to learn novel tasks given a few examples in the form of prompt demonstrations. This paradigm has also been explored on multilingual models with language-specific task descriptions and examples Lin et al., 2022. We leverage this paradigm by prompting various open and closed source LLMs. The models include the above stated along with gemma3 (Team et al., 2025) , gpt4o (OpenAI et al., 2024) and gpt5-mini (Open AI, 2025). The different prompting results for various models is given in Appendix H.

### 2.2 Task-Specific Seq2Seq model

We use ByT5-Sanskrit model (Nehrdich et al., 2024a) as a task-specific Seq2seq model. We hy-pothesie that its domain-specific pretraining on a large Sanskrit corpus provides it with the implicit capability to handle sub-tasks like segmentation and dependency parsing within a unified generation process. We, therefore finetune it directly in a sequence-to-sequence framework.

**ByT5-Sanskrit:** The model is pretrained on a large, multi-domain Sanskrit corpus. The model has achieved state-of-the-art performance on key tasks such as word segmentation, dependency parsing, lemmatization, and morpho-syntactic tagging. Its strong linguistic grounding makes it well-suited for downstream finetuning on poetry-to-prose conversion. We finetune this pretrained model with our two datasets, separately. The model architecture follows that of an encoder-decoder style of a ByT5 model (Xue et al., 2022). To preserve the multitask setup of the base model, we prepend a prefix P2P before each training data following the same setup as that of T0 (Sanh et al., 2022). The schema is shown in Figure 2. The model follows `iast` transliteration and thus all our data is converted to `iast` format.

## 3 Experiment

**Datasets:** To support the task of direct Sanskrit poetry-to-prose conversion, we employ two parallel corpora derived from classical Sanskrit epics: the *Rāmāyaṇa* and the *Mahābhārata*. For the *Rāmāyaṇa* corpus, we extract verse–prose pairs from the Valmiki Ramayana digital repository[4]. The *Mahābhārata* corpus is constructed using parallel data from the Ashramvasika and Ādi Parva sections, sourced from the MAHE Mahabharata portal[5]. Both corpora provide aligned verse–prose sequences, which we use as our supervised data for the conversion task. Table 1 summarises the statistics of the two datasets, comprising a total of 22,860 parallel poetry–prose pairs.

| Dataset | #Train Samples | #Test Samples |
|---|---|---|
| *Rāmāyaṇa* | 16447 | 1829 |
| *Mahābhārata* | 3667 | 917 |

Table 1: Dataset Statistics for *Rāmāyaṇa* and *Mahābhārata* datasets

**Evaluation Metrics:** We use *sacreBLEU* (Post, 2018) with standard tokenization provides a base-line measure of lexical correctness through n-gram overlap; higher scores reflect the presence of key lexical units as well as partial structural alignment captured via higher-order n-grams. Second, we employ *Kendall's Tau*, which quantifies the number of pairwise inversions needed to reorder the model's output into the reference sequence, thereby offering a direct assessment of structural correctness and word-order faithfulness.

**Main Results** We conduct IFT on various multilingual LLMs, and report the results in Table 2. For each of the models, we apply PEFT via the Unsloth library with 4-bit quantization to limit compute. We use LoRA (Hu et al., 2022) with rank `r=16` and `=16`, injecting low-rank adapters in each layer. Prompts follow the `alpaca_prompt` (Taori et al., 2023) and the `chat-template` format augmented with canonical word-ordering rules. Training is performed with TRL's `SFTTrainer` under mixed-precision (`fp16` or `bf16`) to optimize memory and speed.

| Models | *Mahābhārata* | | *Rāmāyaṇa* | |
|---|---|---|---|---|
| | BLEU | KT | BLEU | KT |
| **Instruction Fine Tuning with Rules prompt** | | | | |
| Llama3.1-8B | 28.521 | 0.6260 | 27.564 | 0.6703 |
| Llama3.2-1B | 18.937 | 0.5302 | 16.031 | 0.5362 |
| Llama3.2-3B | 23.308 | 0.5849 | 21.808 | 0.6208 |
| Phi4-14B | 33.123 | 0.6494 | 31.945 | 0.6960 |
| qwen2.5-7B | 25.487 | 0.6209 | 20.467 | 0.6187 |
| **Full-Finetuning** | | | | |
| ByT5-Sanskrit | **38.625** | **0.699** | **39.497** | **0.758** |

Table 2: Results for the IFT experiment on opensourced models compared with the full finetuning of `ByT5-Sanskrit` model.

From Table 2, we observe that FT ByT5-Sanskrit consistently and decisively outperforms all generalist models across both datasets, achieving superior scores on both BLEU and Kendall's Tau.

# 4 Analysis

## 4.1 Can Prompting Replace Task-Specific Fine-Tuning?

In this section, we examine whether LLMs can match the performance of task-specific Seq2Seq models. We evaluate 3 paradigms - Finetuning (FT), Chain-of-Thought prompting (CoT), and Few-Shot prompting with rules (FS-R) and report their BLEU scores in Table 3. Across all configurations, including combinations of FT and CoT, LLMs fail to reach the performance of ByT5-Sanskrit. Finetuning yields the largest gains among LLM approaches, while CoT offers improvements only for weaker models and has limited effect on stronger systems such as GPT-4o. Overall, even with multiple prompting and tuning strategies, LLMs do not match the effectiveness of smaller task-specific Seq2Seq models for this highly structured task.

| Methods | FT | CoT | FS-R | *Mahābhārata* | *Rāmāyaṇa* |
|---|---|---|---|---|---|
| ByT5-Sanskrit | ✓ | - | - | **38.625** | **39.497** |
| Phi4-14B | - | - | ✓ | 11.065 | 7.763 |
| Phi4-14B | - | ✓ | - | 22.572 | 13.771 |
| Phi4-14B | ✓ | - | - | 33.123 | 31.945 |
| Phi4-14B | ✓ | ✓ | - | 30.995 | 24.940 |
| gpt-4o | - | - | ✓ | 24.789 | 20.472 |
| gpt-4o | - | ✓ | - | 24.904 | 21.214 |

Table 3: Fine-tuned ByT5-Sanskrit consistently outperforms all LLM prompting and tuning approaches.

## 4.2 Do Task-Specific Models Generalize Better Than LLMs?

We perform a cross-domain evaluation (Table 4) to assess whether the models learn generalizable syntactic principles or simply memorize training data. We compare the fine-tuned ByT5-Sanskrit model with the best-performing IFT model (Phi4-14B). When trained on the *Mahābhārata* and tested on the *Rāmāyaṇa*, ByT5's BLEU score decreases from 38.63 to 20.34, a notable drop, yet still substantial, indicating effective transfer of Anvaya principles across epic styles. Overall, both models show reduced performance under domain shift, but the nontrivial cross-domain scores demonstrate genuine task learning rather than corpus memorization.

| Model | Mahābhārata | Rāmāyaṇa |
|---|---|---|
| ByT5 finetuned on Mahābhārata | 38.625 | 20.336 |
| ByT5 finetuned on Rāmāyaṇa | 27.420 | 39.497 |
| phi4-14B finetuned on Mahābhārata | 33.123 | 16.072 |
| phi4-14B finetuned on Rāmāyaṇa | 23.001 | 31.945 |

Table 4: Evaluation of BLEU scores for cross-domain analysis on the best performing models

## 4.3 Do Automatic Metrics Align with Human Judgments?

To obtain a gold-standard evaluation, we enlisted a Sanskrit poet to assess 50 outputs (25 from each epic) generated by our best-performing model, the fine-tuned ByT5-Sanskrit. Each output was rated on a 1–10 scale using a weighted scoring scheme described in Appendix F. These human scores were then compared with BLEU and Kendall's Tau, as reported in Table 5. Although the model achieved a moderate BLEU score of 45.82 on the *Mahābhārata* subset, the expert assigned a markedly high average score of 7.74, indicating strong prose quality. Notably, Kendall's Tau exhibited a strong positive correlation with human judgments (0.818), whereas BLEU did not.

| Dataset | BLEU | Kendall's Tau | Human Score (out of 10) |
|---|---|---|---|
| Mahābhārata | 45.82 | 0.818 | 7.74 |
| Rāmāyaṇa | 32.34 | 0.781 | 7.78 |

Table 5: Comparison of human evaluation metric with BLEU and Kendall Tau for a subset of samples from both the datasets. All scores are average scores

## 5 Related Work

Early Sanskrit NLP primarily relied on linguistically grounded, rule-based systems (Huet, 2009; Kulkarni, 2010). Although these approaches captured important grammatical regularities, they faced scalability limitations in handling real-world poetic and prose corpora. With the emergence of annotated datasets (Sandhan et al., 2022a; Krishna et al., 2017, 2020b; Sandhan et al., 2023c), research gradually shifted toward neural, task-specific models. These models span a wide range of applications, including Sanskrit-specific pretraining (Sandhan et al., 2021), Sanskrit ASR (Kumar et al., 2025), compound identification (Krishna et al., 2016; Krishna et al., 2025; Sandhan et al., 2022b), dependency parsing (Sandhan et al., 2023b; Krishna et al., 2020a), segmentation (Nehrdich et al., 2024b; Sandhan et al., 2022c) and word-order linearisation (Krishna et al., 2019, 2020b). Within poetic transformation tasks (Jagadeeshan et al., 2025), early progress in verse-to-prose conversion was achieved using feature-rich Seq2Seq models. These architectures leverage the linguistic structure of Sanskrit, including sandhi rules, compounding templates, dependency relations, and metrical constraints.

Parallel to these developments, general-purpose multilingual and universal LLMs (OpenAI, 2023; Sanh et al., 2022; Abdin et al., 2024) have demonstrated remarkable capabilities under natural-language prompting, instruction tuning, and chain-of-thought reasoning (Kojima et al., 2022; Wei et al., 2022). Their success in English and high-resource languages has motivated attempts to extend LLM-based workflows to low-resource classical languages as well. This contrast raises a central question: with the recent dominance of LLMs, should Sanskrit NLP continue investing in specialised linguistic and neural architectures, or can general-purpose LLMs effectively replace them? Our work directly examines this question for the Sanskrit poetry-to-prose conversion task. We find that despite their breadth, current LLMs still do not match the performance of task-specific encoder–decoder models designed for the linguistic complexities of Sanskrit.

## 6 Conclusion

This work examined whether general-purpose LLMs can replace task-specific models for the structurally demanding task of Sanskrit poetry-to-prose conversion. Our results show that they cannot. Despite advances in instruction-tuning and prompting, LLMs consistently underperform compared to the ByT5-Sanskrit model, which benefits from domain-specific pre-training and full fine-tuning. These findings underscore that, for low-resource and morphologically rich languages, linguistic specialization remains crucial. While prompting strategies grounded in Pāṇinian principles provide a practical alternative when annotated corpora are scarce, they fall well short of the performance achieved through task-specific fine-tuning. Cross-domain evaluation further demonstrates that ByT5-Sanskrit learns generalizable syntactic principles rather than memorizing epic-specific patterns. Finally, strong alignment between Kendall's Tau and expert judgments highlights the importance of structure-aware metrics over surface-level measures like BLEU. Future work should generalize these results to more tasks.

## Limitations

Our study is limited by computational constraints, which restricted closed-API model evaluations to only few advanced reasoning-capable models. The dataset focuses solely on epic poetry from the Rāmāyaṇa and Mahābhārata, limiting generalizability to other Sanskrit genres and meters. We rely on sacreBLEU for evaluation without human judgments, which may not fully capture semantic fidelity or prose fluency. Additionally, our use of transliterated text instead of native Devanāgarī script simplifies processing but overlooks script-specific challenges. We could not release the *Mahābhārata* data publicly due to copy-right regulations.

## Acknowledgements

We would like to express our gratitude to `Swarup Padhi`, `Pretam Ray` (both from IIT Kharagpur), `Harshul Surana` (AI Institute, University of South Carolina), and `Sriram Krishnan` (Central Sanskrit University, Prayagraj) for their help in the initial phases of this work. We are also deeply grateful to `Dr. Arjuna S R` (Assistant Professor - Senior Scale, Manipal Academy of Higher Education, Bengaluru) for the discussions and guidance on the Mahābhārata data, which was invaluable to this study. We also thank `Svarupa`[6], a 501(c)(3) non-profit initiative based in California, USA for providing us 2 x L40 GPUs for this research work. Finally, we sincerely thank the anonymous reviewers for their insightful feedback and constructive criticism, which significantly improved the quality of this paper.

---

# A   Error Analysis for CoT outputs by GPT-4o

Our analysis of GPT-4o's Chain-of-Thought outputs shows that although the model consistently follows the requested step-by-step structure, it often falls short when precise linguistic reasoning is required. The most frequent errors occur in syntactic organization: the relative ordering of cases such as locative and ablative is sometimes incorrect, adjectives in samanādhikaraṇa are misplaced, and clause sequences are occasionally wrongly arranged. Vocatives, which are expected to appear at the beginning of a sentence in standard Sanskrit prose, are sometimes positioned elsewhere. These deviations indicate that the model does not consistently internalize or generalize from the rules and examples provided.

In some cases, the errors originate even earlier, during sandhi resolution. The model occasionally hallucinates morphological forms, such as replacing an ātmanepada verb with its parasmaipada counterpart. Since later CoT steps depend on earlier ones, such initial inaccuracies propagate throughout the reasoning chain and ultimately influence the final prose output. Overall, these patterns suggest that while GPT-4o can reproduce the outward structure of a stepwise explanation, its underlying understanding of Sanskrit grammar and syntax remains uneven, and the explicit CoT format makes these limitations easier to observe.

---

**CoT Examples**

### Example 1

**Input sloka:**
evamuktaḥ sa rājarṣistapovīryasamāśrayat sañcārayāmāsa jarāṃ tadā putre mahātmani
**Ground-truth Prose:**
tadā evam uktaḥ saḥ rājarṣiḥ mahātmani putre tapovīryasamāśrayat jarām sañcārayāmāsa
**Model's CoT Output:**
⟨reasoning⟩
-Applying Rule 1:
evam uktaḥ sa rājarṣiḥ tapovīryasamāśrayat sañcārayāmāsa jarām tadā putre mahātmani
-Applying Rule 2:
Clause 1: sa rājarṣiḥ evam uktaḥ
Clause 2: tapovīryasamāśrayat
Clause 3: tadā sañcārayāmāsa jarām putre mahātmani
-Applying Rule 3:
(evam uktaḥ sa rājarṣiḥ), (tapovīryasamāśrayat) tadā (putre mahātmani jarām)
sañcārayāmāsa
-Applying Rule 4:
(evam uktaḥ sa rājarṣiḥ), (tapovīryasamāśrayat) tadā (mahātmani putre jarām)
sañcārayāmāsa
-Applying Rule 5:
(evam uktaḥ sa rājarṣiḥ), (tapovīryasamāśrayat) tadā (mahātmani putre jarām)
sañcārayāmāsa
⟨/reasoning⟩
⟨answer⟩evam uktaḥ sa rājarṣiḥ tapovīryasamāśrayat tadā mahātmani putre jarām
sañcārayāmāsa⟨/answer⟩
**Issue:** Locative phrase (*mahātmani putre*) incorrectly placed after the ablative (*tapovīryasamāśrayat*), violating Rule 4.3. The temporal adverb *tadā* placed mid-sentence, violating clause-order Rule 2.

### Example 2

**Input sloka:**
kālaḥ pacati bhūtāni kālaḥ saṃharati prajāḥ nirdahantaṃ prajāḥ kālaṃ kālaḥ śamayate
punaḥ
**Ground-truth Prose:**
kālaḥ bhūtāni pacati kālaḥ prajāḥ saṃharati kālaḥ punaḥ prajāḥ nirdahantam kālam
śamayate
**Model's CoT Output:**
⟨reasoning⟩
-Applying Rule 1:
kālaḥ pacati bhūtāni kālaḥ saṃharati prajāḥ nirdahantam prajāḥ kālam kālaḥ śamayati
punaḥ
-Applying Rule 2:

Clause 1: kālaḥ pacati bhūtāni
Clause 2: kālaḥ saṃharati prajāḥ
Clause 3: kālaḥ śamayati punaḥ kālam nirdahantam prajāḥ

```
-Applying Rule 3:
kālaḥ bhūtāni pacati, kālaḥ prajāḥ saṃharati, punaḥ kālam nirdahantam prajāḥ kālaḥ
śamayati
-Applying Rule 4:
kālaḥ bhūtāni pacati, kālaḥ prajāḥ saṃharati, kālaḥ punaḥ prajāḥ nirdahantam kālam
śamayati
-Applying Rule 5:
kālaḥ bhūtāni pacati, kālaḥ prajāḥ saṃharati, kālaḥ punaḥ prajāḥ nirdahantam kālam
śamayati
⟨/reasoning⟩
⟨answer⟩kālaḥ bhūtāni pacati kālaḥ prajāḥ saṃharati kālaḥ punaḥ prajāḥ nirdahantam kālam
śamayati⟨/answer⟩
```
**Issue:** The model hallucinates the parasmaipada form *śamayati* instead of the correct ātmanepada *śamayate*, demonstrating morphological instability.

## B   Appendix: Chain-of-Thought prompt using Anvaya Rules in Alpaca Format

**Chain-of-Thought prompt using Anvaya Rules**

**TASK:** You are a precise prose generator for Sanskrit verses. Given the input Sanskrit sloka, convert it into their canonical Sanskrit prose sentence using the following rules, preserving meaning. Important: The input and output is in IAST transliteration format.

### Canonical Rules for Anvaya Construction:

1. **Sandhi Analysis:** First, resolve all external *sandhi* to isolate individual words. Do not break down integrated compounds (*samāsa*) like 'Ramālayam'.

2. **Clause Structuring:**
   - Identify all clauses (main, subordinate, direct speech).
   - Arrange clauses chronologically based on the events they describe.
   - Relative clauses must precede the main clause they qualify.
   - The finite verb of the main clause concludes the entire sentence.

3. **Modifier Chunking:** Group modifiers with their headwords. These chunks move as a single unit. Place the following directly before the word they describe:
   - Adjectives (*viśeṣaṇa*)
   - Genitives (*ṣaṣṭhī vibhakti*)
   - Adverbs (*kriyāviśeṣaṇa*)
   - Negations (*na*)
   - Appositives (*samānādhikaraṇa*)

4. **Intra-Clause Word Order:** Within each clause, arrange all words and chunks in this precise sequence:
   (a) Vocative (*sambodhana*)
   (b) Subject (*kartṛ*)
   (c) Kārakas (in order): Locative → Ablative → Dative → Instrumental → Accusative
   (d) Non-finite verbs (and their arguments)
   (e) Finite verb (*kriyāpada*) at the very end.

5. **Particle Placement:** Place indeclinable particles (*ca, api, tu, eva, hi*) immediately after the word they modify.

### RESPONSE FORMAT:
Your response should have two different opening and closing tags - ⟨think⟩ ⟨/think⟩ and ⟨prose⟩ ⟨/prose⟩
⟨think⟩A string containing your step-by-step reasoning (the Chain-of-Thought)⟨/think⟩
⟨prose⟩A string containing ONLY the final, Sanskrit prose sentence obtained at the end
of the above thinking process.⟨/prose⟩
Do NOT reveal your internal reasoning beyond the ⟨think⟩ tab or add any commentary - output. Only generate the reasoning explicitly inside ⟨think⟩ and the final prose under ⟨prose⟩.
Given are some examples of sloka and its corresponding prose with its step-by-step reasoning:

### Example 1:
### INPUT:
sloka: *kathaṃ nu rājā vṛddhaḥ sanputraśokasamāhataḥ śokamasmatkṛtaṃ prāpya na mriyeteti cintyate*

**RESPONSE:**

```
⟨think⟩
-Applying Rule 1:
kathaṃ nu rājā vṛddhaḥ san putraśoka-samāhataḥ śokam asmat-kṛtam prāpya na mriyeta iti
cintyate
-Applying Rule 2:
Clause 1: rājā vṛddhaḥ san putraśoka-samāhataḥ
Clause 2: śokam asmat-kṛtam prāpya
Clause 3: kathaṃ nu na mriyeta iti cintyate
rājā vṛddhaḥ san putraśoka-samāhataḥ, śokam asmat-kṛtam prāpya, kathaṃ nu na mriyeta iti
cintyate
-Applying Rule 3:
(putraśoka-samāhataḥ rājā vṛddhaḥ san), śokam asmat-kṛtam prāpya, kathaṃ nu na mriyeta
iti cintyate
-Applying Rule 4:
(putraśoka-samāhataḥ rājā vṛddhaḥ san), asmat-kṛtam śokam prāpya, kathaṃ nu na mriyeta
iti cintyate
-Applying Rule 5:
(putraśoka-samāhataḥ rājā vṛddhaḥ san), asmat-kṛtam śokam prāpya, kathaṃ nu na mriyeta
iti cintyate
⟨/think⟩
⟨prose⟩putraśokasamāhataḥ rājā vṛddhaḥ san asmat kṛtam śokam prāpya kathaṃ nu na mriyeta
iti cintyate⟨/prose⟩
```

**Example 2:**

**INPUT:**

sloka: *ityeṣa sarvabhūtānāṃ mahatāṃ manujādhipa prabhavaḥ kīrtitaḥ samyaṅmayā matimatāṃ vara*

**RESPONSE:**

```
⟨think⟩
-Applying Rule 1:
iti eṣa sarvabhūtānāṃ mahatāṃ manujādhipa prabhavaḥ kīrtitaḥ samyak mayā matimatāṃ vara
-Applying Rule 2:
Clause 1: manujādhipa matimatāṃ vara iti
Clause 2: eṣa sarvabhūtānāṃ mahatāṃ prabhavaḥ kīrtitaḥ samyak mayā
manujādhipa matimatāṃ vara iti eṣa sarvabhūtānāṃ mahatāṃ prabhavaḥ kīrtitaḥ samyak mayā
-Applying Rule 3:
(manujādhipa matimatāṃ vara) iti eṣa (sarvabhūtānāṃ mahatāṃ prabhavaḥ) kīrtitaḥ samyak
mayā
-Applying Rule 4:
(manujādhipa matimatāṃ vara) iti eṣa (sarvabhūtānāṃ mahatāṃ prabhavaḥ) mayā samyak
kīrtitaḥ
-Applying Rule 5:
(manujādhipa matimatāṃ vara) iti eṣa (sarvabhūtānāṃ mahatāṃ prabhavaḥ) mayā samyak
kīrtitaḥ
⟨/think⟩
⟨prose⟩manujādhipa matimatāṃ vara iti eṣa mahatāṃ sarvabhūtānāṃ prabhavaḥ mayā samyak
kīrtitaḥ⟨/prose⟩
```

## C   Few shot Prompt: Canonical Anvaya Generation for Sanskrit Verses

### Few shot prompt using Anvaya Rules

**Task:** You are a precise prose generator for Sanskrit verses. Given the input Sanskrit *śloka*, convert it into their canonical Sanskrit prose sentence using the following rules, preserving meaning.
**Important:** The input and output is in IAST transliteration format.

#### Canonical Rules for Anvaya Construction

1. **Sandhi Analysis:** First, resolve all external *sandhi* to isolate individual words. Do not break down integrated compounds (*samāsa*) like 'Ramālayam'.

2. **Clause Structuring:**
   - Identify all clauses (main, subordinate, direct speech).
   - Arrange clauses chronologically based on the events they describe.

- Relative clauses must precede the main clause they qualify.
- The finite verb of the main clause concludes the entire sentence.

3. **Modifier Chunking:** Group modifiers with their headwords. These chunks move as a single unit. Place the following directly before the word they describe:

   - Adjectives (*viśeṣaṇa*)
   - Genitives (*ṣaṣṭhī vibhakti*)
   - Adverbs (*kriyāviśeṣaṇa*)
   - Negations (*na*)
   - Appositives (*samānādhikaraṇa*)

4. **Intra-Clause Word Order:** Within each clause, arrange all words and chunks in this precise sequence:

   (a) Vocative (*sambodhana*)
   (b) Subject (*kartṛ*)
   (c) *Kārakas* (in order): Locative → Ablative → Dative → Instrumental → Accusative
   (d) Non-finite verbs (and their arguments)
   (e) Finite verb (*kriyāpada*) at the very end.

5. **Particle Placement:** Place indeclinable particles (*ca, api, tu, eva, hi*) immediately after the word they modify.

## Response Format

Your **ENTIRE** response **MUST** be **ONLY** the final Sanskrit prose sentence, enclosed within `<prose>` and `</prose>` tags as shown:

`<prose>This is the final prose sentence</prose>`

## Examples for the Task

**Example 1:**

- **INPUT:**
  sloka: kathaṃ nu rājā vṛddhaḥ sanputraśokasamāhataḥ śokamasmatkṛtaṃ prāpya na mriyeteti cintyate

- **RESPONSE:**
  `<prose>putraśokasamāhataḥ rājā vṛddhaḥ san asmat kṛtam śokam prāpya kathaṃ nu na mriyeta iti cintyate</prose>`

**Example 2:**

- **INPUT:**
  sloka: ityeṣa sarvabhūtānāṃ mahatāṃ manujādhipa prabhavaḥ kīrtitaḥ samyaṅmayā matimatāṃ vara

- **RESPONSE:**
  `<prose>manujādhipa matimatāṃ vara iti eṣa mahatāṃ sarvabhūtānāṃ prabhavaḥ mayā samyak kīrtitaḥ</prose>`

**Example 3:**

- **INPUT:**
  sloka: tatastāpasarūpeṇa prāhiṇotsa bhujaṅgamān phalapatrodakaṃ gṛhya rājñe nāgo'tha takṣakaḥ

- **RESPONSE:**
  `<prose>tataḥ saḥ takṣakaḥ nāgaḥ rājñe phalapatrodakam gṛhya atha bhujaṅgamān tāpasarūpeṇa prāhiṇot</prose>`

# D  Daṇḍa-anvaya-janaka rules

> **Canonical Prose ordering rules from earlier research**
>
> 1. **Sambodhya** (vocative) comes at the initial position in the canonical order.
>
> 2. **Kartṛ** comes after vocative.
>
> 3. **Kāraka** relations follow in reverse order: *adhikaraṇa*, *apādāna*, *sampradāna*, *karaṇa*, and *karman*.
>
> 4. **Viśeṣaṇas**, modifiers with genitive case markers, etc. are placed before their **viśeṣya**.
>
> 5. **Kriyāviśeṣaṇa**, **pratiṣedha**, etc. are placed immediately before their corresponding verb.
>
> 6. **Mukhyakriyā** is positioned at the end of the sentence.
>
> 7. **Avyaya** particles such as *tu* and *api* are placed immediately after their parent word.
>
> 8. The non-finite verbal forms are placed before the **karman**. All the arguments of a non-finite verb appear to their left.
>
> 9. The **kartṛ-samānādhikaraṇa** and **karma-samānādhikaraṇa** are placed after the **kartṛ** and **karman** respectively.

# E  Prompt Ablation Study

We systematically remove each key linguistic instruction (e.g., rules for Sandhi splitting, clause structure, and identifying case endings) from our prompt one by one. We then measured the impact on the model performance for each ablation using BLEU scores on the *Mahābhārata* dataset. The results are shown in Table 6.

| Prompt ablations | BLEU Scores |
|---|---|
| **P_base**: Zero-shot prompt with only task-description | 9.955 |
| **P_full**: The complete 5-rules prompt | 19.584 |
| **P_NoSandhi**: Full prompt minus Rule 1 | 20.005 |
| **P_NoClause**: Full prompt minus Rule 2 | 20.862 |
| **P_NoChunking**: Full prompt minus Rule 3 | 19.633 |
| **P_NoOrder**: Full prompt minus Rule 4 | 20.116 |
| **P_NoParticles**: Full prompt minus Rule 5 | 19.811 |
| **CoT prompt**: Full prompt with intermediate reasoning | 22.572 |

Table 6: Result for the prompt ablation study on different prompts using the Phi4-14B model on the Mahābhārata dataset.

While the linguistic rules outperform the zero-shot baseline (by nearly 100%), the model's performance was highest when not all rules were provided simultaneously. This suggests that there is an optimal balance of instruction, and over-constraining an LLM can be marginally counterproductive. This investigation explores not just the content of the prompt, but its format. The CoT version of our prompt proved to be unequivocally superior, achieving a score of 22.572 - a substantial 15.3% improvement over our original P_full prompt.

# F  Human Evaluation

In this section, we introduce the weighted function used by the Sanskrit poet to score the model predictions from the finetuned ByT5-model. Following are the weighted coefficient for each rule, if being followed. The values are given based on how important that particular rule is with respect to the *anvaya* generation.

1. `Rule 1 = 3`

2. `Rule 2 = 2`

3. Rule 3 = 2

4. Rule 4 = 2

5. Rule 5 = 1

We asked the annotator, who is the co-author of the paper, pursuing his Masters, an expert in Avadhanam[7] with great knowledge on Sanskrit poetry construction and prose understanding.

We report three examples of how the scores are being computed using the outputs of the finetuned Byt5-Sanskrit model in Table 7 .

| Ground truth prose | Model generated | BLEU | Rules followed | Human-scores |
|---|---|---|---|---|
| saḥ sugrīvam mahābalau rāghavau ca praṇamya śūraiḥ vānaraiḥ sahitaḥ divameva utpapāta | saḥ sugrīvam mahābalau rāghavau ca praṇamya śūraiḥ vānaraiḥ sahitaḥ divameva utpapāta ha | 90.360 | 1, 2, 3, 4, 5 | (3+2+2+2+1) = 10 |
| saḥ kurūdvahaḥ brahmadeyā agrahārān ca pradadau tat ca kuntīsutaḥ rājā sarvam eva anvamodata | saḥ kurūdvahaḥ brahmadeyāgrahārān ca pradadau tat ca rājā kuntīsutaḥ sarvam eva anvamodata | 32.774 | 2, 5 | (2+1) = 3 |
| janādhipa tvām anuprāptam manye tat bhavitavyam diṣṭyā śuśrūṣamāṇaḥ tvām manasaḥ jvaram mokṣyāmi | tat janādhipa tvām anuprāptam bhavitavyam manye diṣṭyā śuśrūṣamāṇaḥ tvām manasaḥ jvaram mokṣyāmi | 57.067 | 1, 3, 5 | (3+2+1) = 6 |

Table 7: Scoring methodology for human evaluation

## G IFT on opensource multilingual LLMs

In this study, we demonstrate how the use of *Daṇḍa-anvaya-janaka* (Kulkarni et al., 2019) rules impact the Instruction Following capability of LLMs. We finetune the `Phi4-14B` model with and without using the *Anvaya* rules. We report the BLEU scores for both the datasets.

| Method | Mahābhārata | Rāmāyaṇa |
|---|---|---|
| IFT without rules | 11.234 | 18.905 |
| IFT with rules | 33.123 | 31.945 |

Table 8: Evaluation of *Anvaya* rules on IFT using Phi4-14B model.

Based on the results of Table 8, we observe that providing explicit linguistic rules during Instruction Fine-Tuning (IFT) with the phi4 model leads to a substantial improvement in performance.

## H Prompting Strategies on LLMs

We conduct experiments on the following prompting strategies and compare various open and closed source LLMs to evaluate how the models are performing on the test set of the Mahābhārata dataset and compare the BLEU scores.

1. `Zero-shot without rules` (ZS w/o rules): In this prompt, we only provide the task description without any *Anvaya* rules and examples.

2. `Few-shot without rules` (FS w/o rules): The prompt consists of only the task description and three poetry-prose pairs. It doesn't contain the *Anvaya* rules.

---

3. `Few-shot with rules (FS with rules)`: The prompt consists of task description, custom *Anvaya rules* and along with that three poetry-prose pairs as examples.

4. `Chain-of-Thougt (CoT)`: This prompt consists of our custom step-by-step intermediate reasoning for each *Anvaya* rule with two examples.

| Models | ZS w/o rules | FS w/o rules | FS with rules | CoT |
|---|---|---|---|---|
| **Open-sourced models** | | | | |
| `llama3.1-8B` | 6.458 | - | - | - |
| `gemma3-12B` | 4.342 | 8.276 | 9.767 | 10.596 |
| `phi4-14B` | 8.983 | 10.587 | 11.065 | 22.572 |
| `phi4-14B IFT` | 3.1855 | - | 11.460 | 30.995 |
| **Closed-sourced models** | | | | |
| `gpt4o` | 15.224 | 23.821 | 24.789 | 24.904 |
| `gpt5-mini` | 11.210 | - | 13.778 | 14.071 |

Table 9: BLEU scores for different prompting strategies on both open and closed sourced models on the Mahābhārata dataset

We observe that the BLEU scores increases when we provide the linguistic rules as compared to giving only task description with and without examples. The Chain-of-Thought prompting is highly beneficial for the open-sourced models, specially on finetuned models which scored even higher than the frontier models such as gpt4o and gpt5-mini.